\documentclass[runningheads]{llncs}
\usepackage[T1]{fontenc}
\usepackage{graphicx}
\usepackage{subcaption}
\usepackage{tikz}
\usepackage{amsmath}
\usepackage{array}
\newcolumntype{P}[1]{>{\centering\arraybackslash}p{#1}}
\usepackage{amsfonts}
\usepackage{bbold}

\newcommand{\R}{\mathbb{R}}
\newcommand{\un}{\mathbb{1}}

\newcommand{\ud}{\mathrm{d}}

\usepackage{url}
\usepackage[bottom]{footmisc}

\begin{document}
\title{Fast Marching Energy CNN}
%
%
\author{Théo Bertrand$^*$\inst{1} \and Nicolas Makaroff$^*$\inst{1} \and Laurent D. Cohen\inst{1}}
\authorrunning{T. Bertrand, N. Makaroff et al.}

%
\institute{CEREMADE, UMR CNRS 7534, University Paris Dauphine, PSL Research University, 75775 Paris, France
\email{\{bertrand, makaroff, cohen\}@ceremade.dauphine.fr}\\
}
\maketitle              
\begin{abstract}
Leveraging geodesic distances and the geometrical information they convey is key for many data-oriented applications in imaging. Geodesic distance computation has been used for long for image segmentation using Image based metrics.
We introduce a new method by generating isotropic Riemannian metrics adapted to a problem using CNN and give as illustrations an example of application. We then apply this idea to the segmentation of brain tumours as unit balls for the geodesic distance computed with the metric potential output by a CNN, thus imposing geometrical and topological constraints on the output mask. We show that geodesic distance modules work well in machine learning frameworks and can be used to achieve state-of-the-art performances while ensuring geometrical and/or topological properties. 
\end{abstract}
\section{Introduction}
\let\thefootnote\relax\footnotetext{$^*$Equal contribution}

Geodesic curves and distances have been used to convey geometric properties in many different applications. The usual approach of those methods is to rely on prior knowledge of the task at hand to build a Riemannian metric $g$ explicitly from data.

The approach presented in this work tries to get rid of the bias introduced in the choice of a metric tensor by generating it from data via a Neural Network architecture which parameters were previously optimized in a supervised learning approach with training data. Introducing such a bias is not a bad thing in itself, however, it requires an arbitrary decision from a user and parameter tuning, two issues that can be avoided by learning to generate a metric from data.

To demonstrate the effectiveness of this framework, we apply it to a segmentation task using a brain tumour MRI images dataset. By using our proposed method, we can obtain accurate results compared to traditional approaches, highlighting the capabilities of this approach. Furthermore, we also observe that our method has a remarkable ability to learn from data and somewhat generalize to unseen data.

The method introduced in this work offers a powerful and flexible way of using geodesic curves and distances in a wide range of applications in a holistic learning framework.

The rest of the paper is organised as follows. In Section \ref{sec:comp_geod} we present the computation of geodesic distances and their gradient. In Section \ref{sec:model} we introduce our experimental method for Fast Marching Energy CNN. In Section \ref{sec:experiment} we present the main results of our experiments and provide a discussion around our work.
\subsection*{Related Works}
The use of geodesic distances in segmentation tasks has a long history. To the authors' knowledge, the first article to segment an image's region using a minimal path distance and fast marching is \cite{malladi_sethian}, with application on a 3D brain image. In the case of the segmentation of tubular tasks we can refer to \cite{chen:hal-01415036} for instance, a method that segments the 3D vascular tree by propagating the front of the minimal path distance computation. Similarly, \cite{COhen_Deschamps} segments vascular structures by introducing an anisotropic metric, determined dynamically by evaluating local orientation scores during the Fast Marching computations. Those 3 articles already use the level sets of the geodesic distance (or "geodesic balls") to provide the segmentation mask. We may also mention \cite{benmansour_cohen_tubular} that uses geodesic curves in an higher dimensional space to track vessels (as curves with an additional width component).
These works generally aren't interested in treating the task in an holistic manner and focus on providing a good model for the structures to segment, whereas this work tries to treat the problem end-to-end and generalize to a large dataset of input images.

Only a few previous methods are interested in learning a metric from data. 
We may mention recent works such as \cite{Scarvelis_solomon_2022} and \cite{heitz:hal-02989081} that try to find metric tensors that fit spatio-temporal data in order to capture the velocity fields and underlying geometry of the data. The first paper is modelling trajectories as the solutions of a dynamical system generated by a Neural Network and also taking into account the dynamics of the whole population by penalizing an optimal transport cost between two consecutive timestamps. However \cite{heitz:hal-02989081} tries to interpolate a sequence of histograms with Wasserstein barycenters by optimizing over the metric tensor appearing in the ground cost. Also, there are important links between the Wasserstein optimal transport, its dynamical formulation and geodesics, for further reading, we refer to \cite{ambrosio2021lectures}.
These works propose interesting frameworks to work with, but they are not focused on generalizing the generation of the metric tensors.


\cite{benmansour:hal-00360794} is an older article that is important for our work, as they laid the ground for the differentiation of the geodesic distance with respect to the metric in the Fast Marching algorithm. They then proceed to apply it in the setting of inverse problems to retrieve the metric from distance measurements. Its only concerns were to solve inverse problems involving the geodesic distance, whereas we go one step further by including a Fast Marching module in a deep learning segmentation procedure. The sub-gradient marching algorithm is briefly described in section 2 as it is essential to our framework to propagate through the Fast Marching module and carry the learning step.

In terms of Deep Learning, we might add a few references such as the classical \cite{unet} and \cite{resnet} that respectively introduce the UNet and ResNet architectures, which are used for our method and as baseline comparisons. For a review of deep learning methods in medical imaging one might refer to \cite{review_zhou_2021}. The very general methods directly producing segmentation from medical images are already quite efficient, but they suffer from a lack of robustness and do not impose a lot of structure on the segmentation that comes out of the network. Contrary to this, our work allows to impose a lot of constraint on the topology of the segmented region (namely a set with trivial topology).

\section{Computing geodesic distances and their gradient}\label{sec:comp_geod}

The geodesic distance is a fundamental concept in the field of Riemannian geometry, and it is used to quantify the distance between two points on a (compact, path-connected) manifold $\mathcal{M}$. It is defined as the minimal length of all possible paths linking two points on the manifold.

Formally, the geodesic distance is given by the following :
\begin{equation}
d_g(x,y) = \inf_{\gamma\in \mathrm{Lip}([0,1], \mathcal{M}), \gamma(0)=x, \gamma(1)=y}\int_0^1 \sqrt{g_{\gamma(t)}(\gamma'(t),\gamma'(t))} \ud t, 
\end{equation}
where $ \mathrm{Lip}([0,1], \mathcal{M})$ is the space of Lipschitz curves on the manifold $\mathcal{M}$ and parameterized by the interval $[0,1]$. $g$ is a metric tensor, which is a map defined at each point $x\in \mathcal{M}$ as $g_x : (u,v) \in \mathcal{T}_x \mathcal{M}^2 \mapsto g_x(u,v)$ is positive definite bilinear form. This means that $\sqrt{g_x}$ is a Euclidean norm on $\mathcal{T}_x\mathcal{M}$, the tangent space to $\mathcal{M}$ at point $x$.

In this work, we will consider a very simple mathematical framework, where $\mathcal{M}$ is simply a path-connected, open and bounded set $\Omega$ of $\R^d$ and $\mathcal{T}_x\mathcal{M}$ can be identified with $\R^d$. This simplification allows for a more straightforward implementation of the geodesic distance, while still maintaining its core properties and mathematical foundation. In the following we will have $g_x(u,v) = \phi(x)^2 \left<u,v\right>_{\R^d}.$

\subsection*{Fast Marching Algorithm}
Since the seminal work of \cite{sethian_1996}, the Fast Marching algorithm has been one of the most widely used methods for computing geodesic distances on a manifold. The Fast Marching method computes the geodesic distance by front propagation.

The Eikonal equation, which has the geodesic distance as its unique positive viscosity solution, is the key component to the front propagation in Fast Marching.

The distance $u$ from a set $S\subset \Omega$ satisfies the Eikonal equation: 


\begin{equation}
     \left\{
            \begin{array}{ll}
                \forall x \in \Omega \setminus S, & \|\nabla u(x)\| = \phi(x),\\
              \forall x\in S, & u(x) = 0,
            \end{array}
          \right.
          \label{eq:Eikonale}
\end{equation}
\noindent
It can be shown that the unique positive solution to the equation \eqref{eq:Eikonale} in the sense of viscosity solutions is the geodesic distance from the set $S$, relative to the metric tensor field associated with the matrices $\phi(x)^2 \mathbf{I}_d$.

The Eikonal equation is discretized using the upwind scheme :
\begin{equation}
    \sum_{1\leq i\leq 2} \frac{1}{h^2} \max(u_p-u_{p+e_i},u_p-u_{p-e_i},0)^2  = \phi_p^2,
    \label{eq:upwind_scheme}
\end{equation}
\noindent
with $u_p$ and $\phi_p$ the geodesic distance and potential at point $p$ in the discretized domain $\Omega$, $p\pm e_i$ denote the adjacent points on the grid and $h$ is the discretization parameter.

Fast Marching is an algorithm that iteratively visits each point on the grid from neighbour to neighbour. At each iteration we look at the neighbour points to those that have already been Accepted, and we accept the nearest point among the neighbours and we repeat by computing the new neighbourhood of the Accepted points. We initialise all values at $+\infty$ except the seed point at $0$. Depending on the number of accepted points connected to $p$ on the grid, equation \eqref{eq:upwind_scheme} reduces either to a quadratic of affine equation to find $u_p$ from the values of the parent points.

In practice, we use the python library \textit{Hamiltonian Fast Marching} (\textit{HFM}) that provides a fast and efficient implementation of the Fast Marching method and of the so-called Subgradient Marching Algorithm \cite{mirebeau_hfm}.

\subsubsection*{Differentiating Fast Marching}

The ability to differentiate the geodesic distance with respect to the metric is an important tool in many applications, such as shape optimization and optimal control. The first work to propose a numerical method to differentiate the geodesic distance with respect to the metric is \cite{benmansour:hal-00360794}, and it has found few applications (see for instance \cite{bonnivard:hal-01791129}).

To differentiate the geodesic distance, we can use the update in the discretized Eikonal equation \eqref{eq:discrete_eikonal}. By taking the Eikonal equation written in dimension 2, and using the setting of interest, i.e. an isotropic metric $g_x(v,w) = \phi(x)^2 \left<v,w\right>_{\R^2},$ we write the discretized version of the Eikonal equation, with $h$ the discretization parameter, the discretized domain is simply a regular square grid :
\begin{equation}
    \begin{cases}
        (u_p-u_{p\pm e_1})^2 + (u_p-u_{p\pm e_2})^2 = h^2 \phi_p^2 \text{ if p has 2 parents},\\
        u_p = \min_iu_{p\pm e_i} + h \phi_p\text{ if p has only 1 parent or $h^2\phi_p^2 < (u_{p\pm e_1} - u_{p\pm e_2})^2$}.       
    \end{cases}
    \label{eq:discrete_eikonal}
\end{equation}
\noindent
Thus $u_p$ is the value of the distance computed by fast marching at point $p$, and we define $D_{\phi}u_p\in \R^{n^2}$ the differential of $u_p$ with respect to the potential $\phi$.

Differentiating with respect to $\phi$ in the two cases of update, we get
\begin{multline}
    \begin{cases}
        (u_p - u_{p\pm e_1}) (D_{\phi}u_p- D_{\phi}u_{p\pm e_1})  + (u_p - u_{p\pm e_2}) (D_{\phi}u_p- D_{\phi}u_{p\pm e_2})  = h^2 \phi_p \\ \text{ if p has 2 parents},\\
        D_{\phi}u_p  = D_{\phi} u_{p\pm e_i} + h\un_{p} \text{ if p has only 1 parent or $h^2\phi_p^2 < (u_{p\pm e_1} - u_{p\pm e_2})^2$},
    \end{cases}
\end{multline}
\noindent
with $\un_p \in \R^{n^2}$ the vector filled with zero except at coordinate $p,$ which gives the update:
\begin{equation}
    \begin{cases}
        D_{\phi}u_p = \frac{(u_p - u_{p\pm e_1}) D_{\phi}u_{p\pm he_1}  + (u_p - u_{p\pm e_2}) D_{\phi}u_{p\pm e_2}  + h^2 \phi_p}{(u_p - u_{p\pm e_1}) + (u_p - u_{p\pm e_2})}\text{ if p has 2 parents},\\
        D_{\phi}u_p =  D_{\phi}u_{p\pm e_i} + h\mathbb{1}_{p} \text{ if p has only 1 parent or $h^2\phi_p^2 < (u_{p\pm e_1} - u_{p\pm h_2})^2$},
    \end{cases}
\end{equation}
\noindent
This update can then be used to compute the gradient of the geodesic distance with respect to the metric tensor during the Fast Marching iterations. In \cite{benmansour:hal-00360794} it is named \textit{Subgradient Marching Algorithm}. This method can be extended to higher dimensions as well as more general Finsler metrics.

\section{Model}\label{sec:model}

\begin{figure}[ht]
\includegraphics[width=\textwidth]{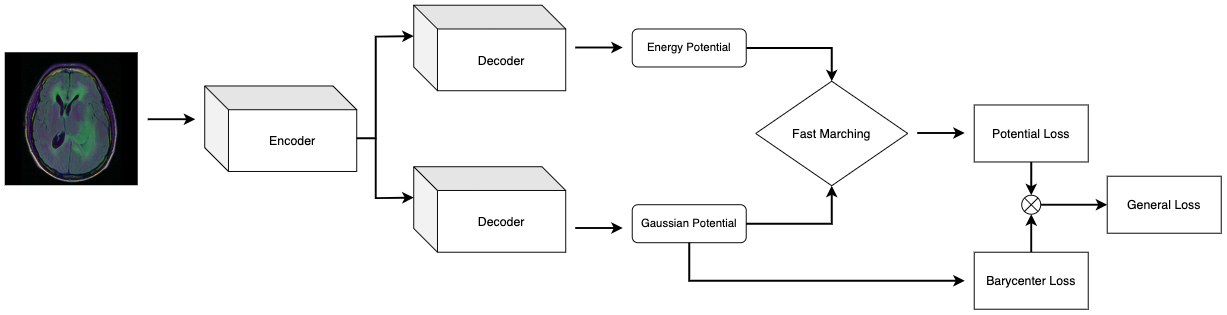}
\caption{Diagram of the framework from the input image to the loss.}
\label{fig:diagram}
\end{figure}

The proposed method presented in this study uses a neural network, specifically a modified version of the UNet architecture, to segment regions of an image as geodesic balls with respect to a metric. The metric is obtained by training a convolutional neural network (CNN) to provide both the metric and the center or seed of the geodesic ball. The framework, as shown in Figure \ref{fig:diagram}, processes the input image using the encoder component of the UNet, resulting in a vector representation of the image. This vector is then passed through two separate decoders to perform distinct tasks.

The first decoder predicts the potential $\phi$ to be used by the fast marching module, which can be computed using the HFM library. The second decoder predicts a Gaussian potential that represents the probability of the presence of the region's barycenter in a given area, which is also provided as a seed to the fast marching module. The distance map generated by the fast marching procedure is then used to find a geodesic ball for segmentation. The expected segmentation is compared to the predicted segmentation, and the theoretical barycenter is compared to the predicted Gaussian potential to compute the error. 

The distance computation module can be written as a function of both seed points and input metric. The metric $\phi$ is defined as the output of a CNN architecture, such as the widely used UNet, with $\theta$ being in the space of parameters. We enforce positive and non-zero properties of the metric by taking $\phi = f_{\theta}(u)^2 + \epsilon$, with $u$ being the input image and let $f_{\theta}$ be a CNN, with $\theta\in\R^p$ the space of parameters. To avoid solutions that distribute a lot of mass everywhere, as noted in \cite{benmansour:hal-00360794}, we ensure that the total mass of the metric is reasonable by applying a transformation $\phi \mapsto \frac{\phi}{\max(\frac{1}{\lambda}\|\phi\|_1,1)}$ that upper bounds the $L^1$ norm at a fixed level $\lambda$ (We chose in this work to empirically bound the total mass at 5).

\subsection*{UNet}

In this study, we focus on the task of potential generation and employ two different architectures commonly used for image segmentation: the UNet \cite{unet} and a combination of the UNet and ResNet \cite{resnet}. The UNet is a fully convolutional neural network that is designed for image segmentation, comprising of a contracting path and an expansive path. The contracting path reduces the spatial resolution of feature maps while the expansive path increases it. The combination of these paths allows for the extraction of high-level features from the input image and recovery of the spatial resolution to provide a segmented output.

However, the depth of CNNs can cause the problem of vanishing gradients, which can affect model performance. To address this, we propose the use of ResNet-UNet, a combination of the UNet and ResNet-34 model in the encoder portion of the network. ResNet-34 benefits from deep residual learning and comprises of a 7x7 convolutional layer, a max pooling layer, and 16 residual blocks.

By combining these architectures, ResNet-UNet can capture fine and coarse features of input images and learn deeper and more complex representations. This results in a more accurate and robust model for image segmentation tasks, as demonstrated by our experimental results. Additionally, we introduced modifications to the expansive path of both networks, implementing a dual expansive path system to predict potential energy and a Gaussian potential for the prediction of barycenter. These modifications are illustrated in Figure \ref{fig:diagram}. Overall, our proposed model demonstrates promising results for potential generation tasks.

\subsection*{Generating masks with geodesic balls}

Applications may take advantage of topological priors on the label to reconstruct. For instance one may need to recover regions in an image that we know to be path-connected and of trivial topology. Such regions might be modelled as balls related to a specific distance and recovered as indicator function of such a ball. Formally, we expect for a set $E$ to recover an indicator function as $\chi_{d_{\phi}(x_0,\cdot)\leq 1}$ for well chosen $x_0 \in \R^d$ and $\phi \in L^1(\Omega).$

With this method of building masks for specific tasks, we can try to generalize using a neural network architecture and find good potential $\phi$ to segment interesting regions in images. To do this we would need to compute the gradient of a chosen loss function and thus would need to differentiate the mask, that is why we will replace the indicator function on the unit ball, that would yield zero gradients almost everywhere, by a sigmoid that will smoothly interpolate between the value $1$ in the region inside the unit ball and $0$ outside. Given the distance map $d_{\phi}(x_0,\cdot)$, our mask then becomes $\chi^{\delta}(d_{\phi}(x_0,\cdot)) = 1 - \frac{1}{1+ \exp(-(d_{\phi}(x_0,\cdot)-1)/\delta)},$ which approaches characteristic function of the unit ball as the parameter $\delta$ approaches $0.$ $\delta$ will be taken typically of the order of the size of pixel, i.e. approximately the inverse of the image size.

\begin{figure}
    \begin{subfigure}{0.5\textwidth}
        \includegraphics[width=\textwidth, trim=0 3cm 0 3cm, clip]{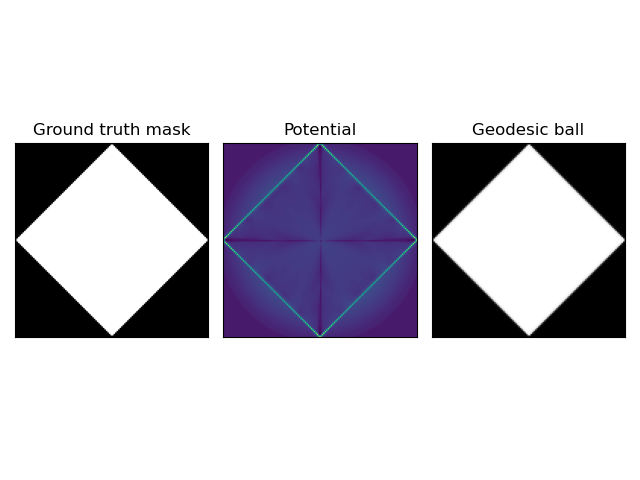}
    \end{subfigure}
    \hspace{0.01cm}
    \begin{subfigure}{0.5\textwidth}
        \includegraphics[width=\textwidth, trim=0 3cm 0 3cm, clip]{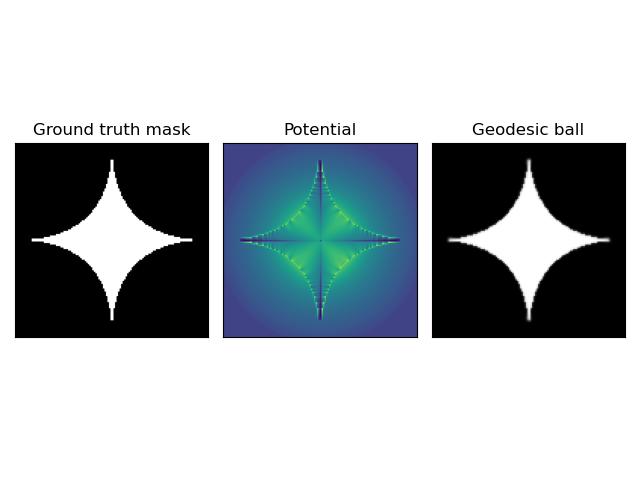}
    \end{subfigure}
    
    \caption{Example of recovery of an isotropic metric fitting two regions by minimizing $\|\chi^{\delta} \circ d_{\phi^2} - y\|^2_2$ with respect to $\phi$, where $y$ is the ground truth mask, $\delta = 0.01$. $x_0$ is taken as the center of the mask to be recovered.}
    \label{fig:fitting_balls}
\end{figure}

Figure \ref{fig:fitting_balls} shows how it is possible to approach the characteristic function of different sets with this formulation. This problem is not convex, so solutions may vary depending on the initialization for instance, but it seems that most of the time potentials converge to a solution that puts a lot of mass on the edges of the mask to recover. The seed here is fixed to $x_0$ the center of the balls to be fitted, and the potential $\phi$ is directly optimized using automatic differentiation and ADAM with a "learning rate" equal to $0.01$. $\phi^2$ is taken as input for the fast marching algorithm instead of $\phi$ as an easy way to smoothly enforce positivity of the potential.

\section{Experiments}\label{sec:experiment}
As announced before, our experiments were led on tumour segmentation task. 

\subsection*{Data} 
To conduct our experiments we have used a dataset of Brain MRI segmentation task that is the TCGA\_LGG database openly available on the internet \cite{tcga_lgg}. This database contains MRI scans of patients with brain tumours. They correspond to 110 patients (resulting in 1189 images) included in The Cancer Genome Atlas (TCGA) lower-grade glioma collection with at least fluid-attenuated inversion recovery (FLAIR) sequence and genomic cluster data available. We removed tumour with multiple connected components. This dataset is composed of the data of 110 patients. 
We have used the set of 2D MRI images as our learning and training datasets. We have set aside a 10 patients' data to form a test set as independent as possible (whereas two images from the same patient can be separated in the training and validation set, test data are always the result of a different acquisition from the training and validation set).
We applied data augmentation on the training images to increase the diversity of the training set and improve the generalization of the model. The data augmentation techniques used were: horizontal flipping with probability p=0.5, vertical flipping with probability p=0.5, random 90-degree rotation with probability p=0.5, transpose with probability p=0.5, and a combination of shifting, scaling, and rotating with probability p=0.25. We respectively set the shift limit, scale limit and rotation limit to 0.01, 0.04, and 0 (as we already perform rotation). 
We computed the tumour seed using a Euclidean barycenter of the mask region.

\subsection*{Model Training Procedures}

The UNet architecture was employed for the task of image segmentation in this study. The model was initialized with Kaiming distribution and trained using the Adam optimizer, which has been widely used in literature due to its capability to adjust the learning rate during training. The learning rate was set to 1e-3, which is a commonly used value in CNNs, as it provides a balance between achieving convergence and avoiding overshooting the optimal solution. In order to optimize the model's performance, to penalize the error between the prediction mask and the groundtruth mask we used a combination of Dice loss and Binary Cross-Entropy (BCE) loss (Equation \eqref{eq:loss}).
\begin{equation}
    \mathcal{L}_{S}(x,y) = \frac{2\times \sum_{i=1}^N x_i y_i}{\sum_{i=1}^N x_i + \sum_{i=1}^N y_i} + \frac{1}{N}\sum_{i=1}^N -(x_i\log(x_i) + (1-y_i)\log(1-y_i))
    \label{eq:loss}
\end{equation}
To control the error on the seed prediction a Binary Cross-entropy loss was used. 
\begin{equation}
    \mathcal{L}_{H}(h^1,h^2) = \frac{1}{N}\sum_{i=1}^N -(h^1_i\log(h^1_i) + (1-h^2_i)\log(1-h^2_i))
\end{equation}
The final loss is:
\begin{equation}
    \mathcal{L}(x,y,h^1,h^2) = \mathcal{L}_{S}(x,y) + {L}_{H}(h^1,h^2)
\end{equation}
The Dice loss function, which is known for its ability to handle imbalanced data, was combined with the BCE loss function, which provides stability during training. 

In order to determine the distance between two barycenters, a transformation of the position coordinates into a Gaussian potential is used, based on the following formulation:
\begin{equation}
f(x,y) = \frac{1}{\sqrt{2\pi}\sigma}\exp(\frac{(x - b_i)^2 + (y-b_j)^2}{2\sigma^2})
\end{equation}
Here, $(b_i, b_j)$ represent the coordinates of the barycenter. At inference time, the predicted potential is used to identify the maximum location, from which the barycenter coordinates can be extracted.

The model's architecture was initialized with 64 feature maps, which has been shown to be a suitable number for high resolution images, and a batch size of 16 was used during the training process. This combination of hyperparameters allowed the model to effectively use detailed information from the input image while maintaining a balance between generalization and overfitting, as demonstrated by the results presented in this paper. Perhaps it should be clarified that since the two decoders are different and predict two different things, these new parameters do not assist the segmentation compared to the direct method.

\subsection*{Potential Analysis}
\begin{figure}
    \centering
    \begin{subfigure}{0.18\textwidth}
        \rotatebox{90}{\includegraphics[width=\textwidth]{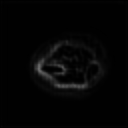}}
        \subcaption{Epoch 1}
    \end{subfigure}
    \hspace{0.01cm}
    \begin{subfigure}{0.18\textwidth}
        \rotatebox[origin=c]{180}{{\includegraphics[width=\textwidth]{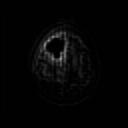}}}
        \subcaption{Epoch 10}
    \end{subfigure}
    \hspace{0.01cm}
    \begin{subfigure}{0.18\textwidth}
        \rotatebox[origin=c]{270}{\includegraphics[width=\textwidth]{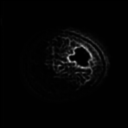}}
        \subcaption{Epoch 50}
    \end{subfigure}
    \hspace{0.01cm}
    \begin{subfigure}{0.18\textwidth}
        \rotatebox[origin=c]{90}{\scalebox{-1}[1]{\includegraphics[width=\textwidth]{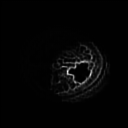}}}
        \subcaption{Epoch 100}
    \end{subfigure}
    \hspace{0.01cm}
    \begin{subfigure}{0.18\textwidth}
        \scalebox{-1}[1]{\includegraphics[width=\textwidth]{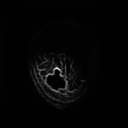}}
        \subcaption{Epoch 150}
    \end{subfigure}
    \caption{Evolution of the predicted potential taken as input in the Fast Marching Module.}
    \label{fig:pot_epoch}
\end{figure}

 The potential generated by the neural network was analyzed with respect to the number of training epochs. Results show on Figure \ref{fig:pot_epoch} that the output distribution quickly converged towards the boundaries of the tumour to be segmented. However, as training progressed, the contour of the tumour sharpened and boundaries became more distinct and at the same time we can see the brain edges removed. The potential in the end only holds detailed information of the contours in a small area around the tumour.

 \subsection*{Segmentation Experiments}

\begin{figure}[ht]
\centering
\begin{subfigure}{0.23\textwidth}
\includegraphics[width=\textwidth]{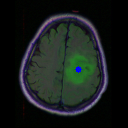}
\end{subfigure}
\hspace{0.01cm}
\begin{subfigure}{0.23\textwidth}
\includegraphics[width=\textwidth]{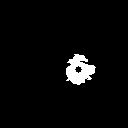}
\end{subfigure}
\hspace{0.01cm}
\begin{subfigure}{0.23\textwidth}
\includegraphics[width=\textwidth]{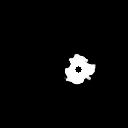}
\end{subfigure}
\hspace{0.01cm}
\begin{subfigure}{0.23\textwidth}
\includegraphics[width=\textwidth]{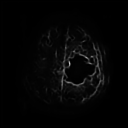}
\end{subfigure}


\begin{subfigure}{0.23\textwidth}
\includegraphics[width=\textwidth]{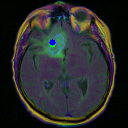}
\caption{Input Image}
\end{subfigure}
\hspace{0.01cm}
\begin{subfigure}{0.23\textwidth}
\includegraphics[width=\textwidth]{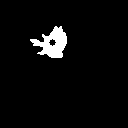}
\caption{Groundtruth}
\end{subfigure}
\hspace{0.01cm}
\begin{subfigure}{0.23\textwidth}
\includegraphics[width=\textwidth]{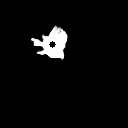}
\caption{Predicted}
\end{subfigure}
\hspace{0.01cm}
\begin{subfigure}{0.23\textwidth}
\includegraphics[width=\textwidth]{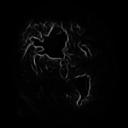}
\caption{Potential}
\end{subfigure}

\caption{Results of the segmentation on validation data. On the input image, the blue and green dots are respectively the groundtruth and predicted seed.}
\label{fig:prediction}
\end{figure}

 We compared our method to a standard UNet segmentation approach. As can be seen in the results plots \ref{fig:prediction}, our method demonstrates clear edge detection. The well-defined contours produced by our method are a result of its ability to take into account the morphology of the image, which traditional filters are not able to do. Furthermore, the problem-specific nature of our method allows for improved performance in image segmentation. Classical metrics allows us to compare quantitatively the results of our segmentation. Overall we recover the same precision on the segmentation mask with minimal improvements of the symmetric Hausdorff distance. However the convergence towards an acceptable solution is faster when combined with the Fast Marching Module since with only a approximate potential the method converge to a relatively close segmentation. Time gives the neural network to more precisely learn the filter and sharpens the edge of the tumour. A general observation from the segmentation in Figure \ref{fig:prediction} is that the method when failing to predict correctly a pixel tends to create false positive rather than true false. The Table \ref{tab:results} shows how our method has a high recall controlling that there is a very low number of false negative. We performed the training with the library \textit{HFM} and the heat method and recorded same results. Overall the UNet architecture shows difficulties to precisely learn the potential while from a metric point of view the ResNet-UNet performs comparatively as the classical segmentation technique using CNNs. 

\begin{table}
\centering
\caption{Segmentation results (IOU) on the TGCA\_LGG brain MRI database.}
\begin{tabular}{|p{10.8em}|P{4em}|P{4em}|P{4.2em}|P{4em}|P{4em}|P{4em}|}
\hline
Name &  Dice & IOU & Hausdorff & F1 Score & FPR & FNR  \\
\hline
UNet & $0.862 $ & $0.869 $ & $2.313 $ & $0.869 $ & $0.007$& $0.05$ \\
ResNet UNet & $0.873 $ & $0.877 $ & $2.257 $ & $0.877$ & $0.006$ & $0.07$\\
FM UNet (ours)& $0.825 $ & $0.823 $ & $2.505 $ & $0.823$& $0.011$ &$0.064$\\
FM Resnet UNet (ours)& $0.863 $ & $0.866 $ & $2.248 $ & $0.866$ & $0.009$& $0.04$\\
\hline
\end{tabular}
\label{tab:results}
\end{table}

\begin{figure}[ht]
\centering
\begin{subfigure}{0.23\textwidth}
\includegraphics[trim={3cm 1cm 3cm 1cm},clip,width=\textwidth]{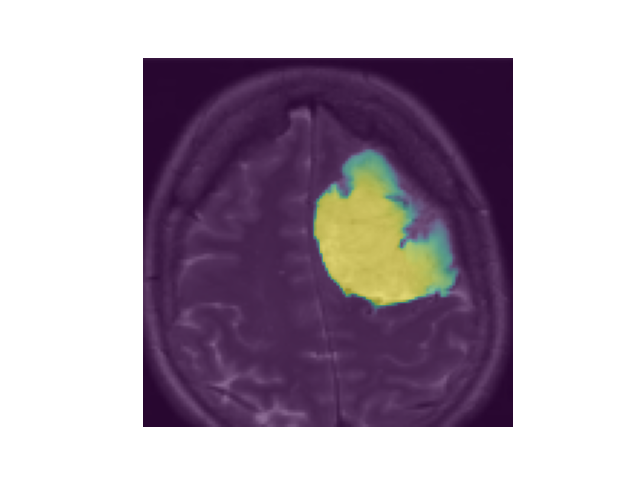}
\end{subfigure}
\hspace{0.01cm}
\begin{subfigure}{0.23\textwidth}
\includegraphics[trim={3cm 1cm 3cm 1cm},clip, width=\textwidth]{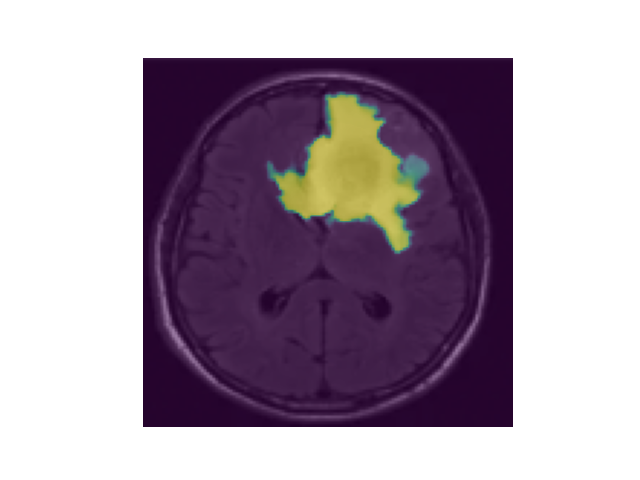}
\end{subfigure}
\hspace{0.01cm}
\begin{subfigure}{0.23\textwidth}
\includegraphics[trim={3cm 1cm 3cm 1cm},clip, width=\textwidth]{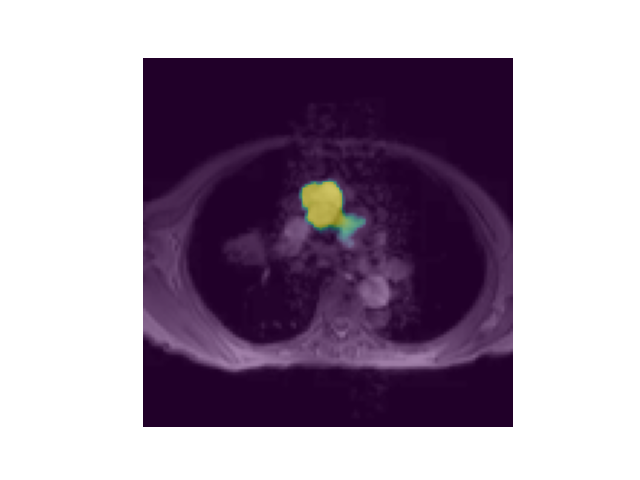}
\end{subfigure}
\hspace{0.01cm}
\begin{subfigure}{0.23\textwidth}
\includegraphics[trim={3cm 1cm 3cm 1cm},clip, width=\textwidth]{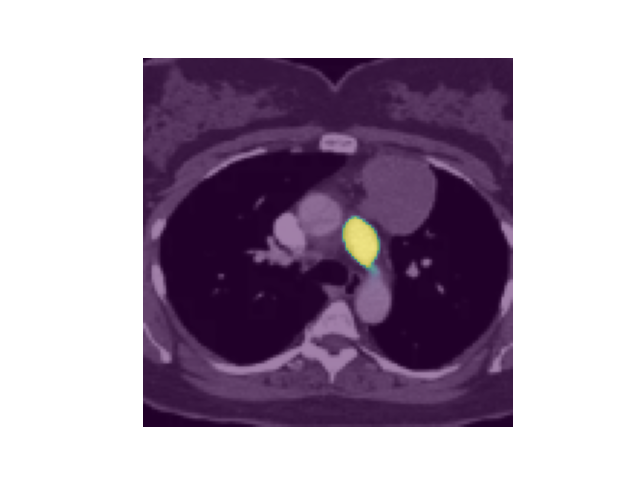}
\end{subfigure}

\begin{subfigure}{0.23\textwidth}
\includegraphics[trim={3cm 1cm 3cm 1cm},clip,width=\textwidth]{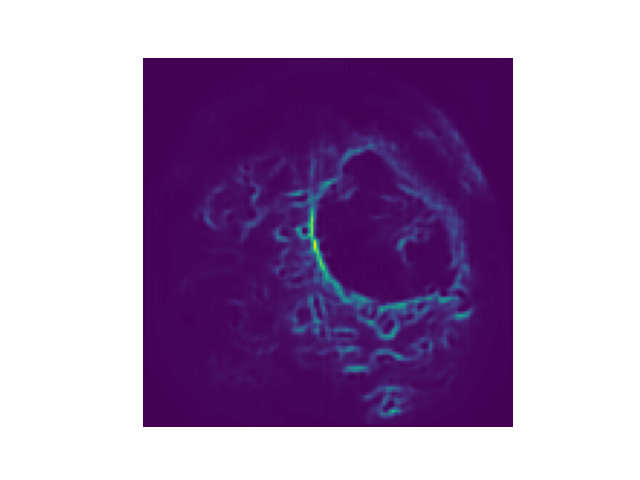}
\end{subfigure}
\hspace{0.01cm}
\begin{subfigure}{0.23\textwidth}
\includegraphics[trim={3cm 1cm 3cm 1cm},clip,width=\textwidth]{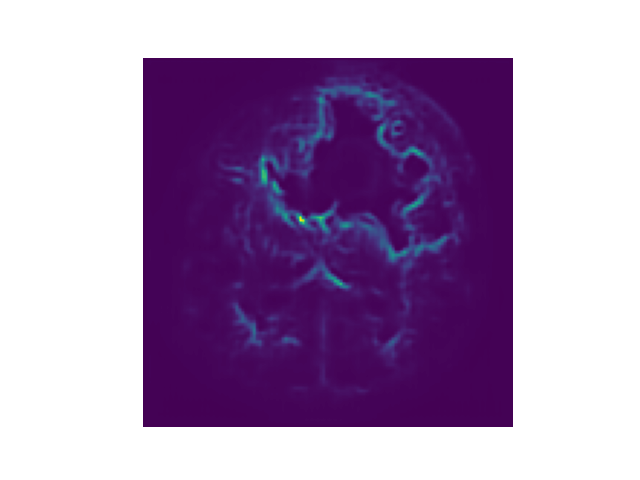}
\end{subfigure}
\hspace{0.01cm}
\begin{subfigure}{0.23\textwidth}
\includegraphics[trim={3cm 1cm 3cm 1cm},clip,width=\textwidth]{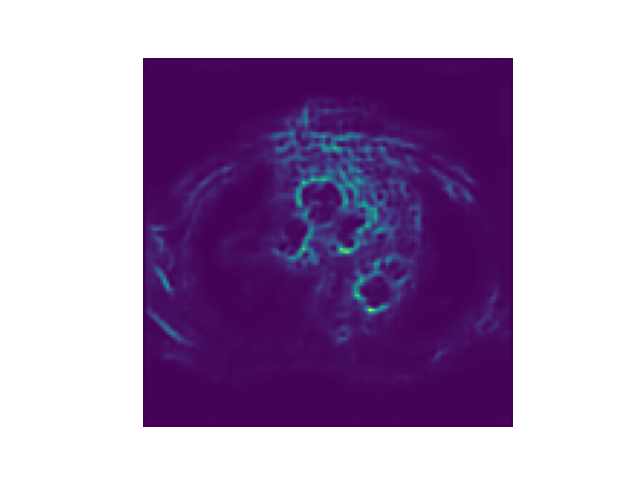}
\end{subfigure}
\hspace{0.01cm}
\begin{subfigure}{0.23\textwidth}
\includegraphics[trim={3cm 1cm 3cm 1cm},clip,width=\textwidth]{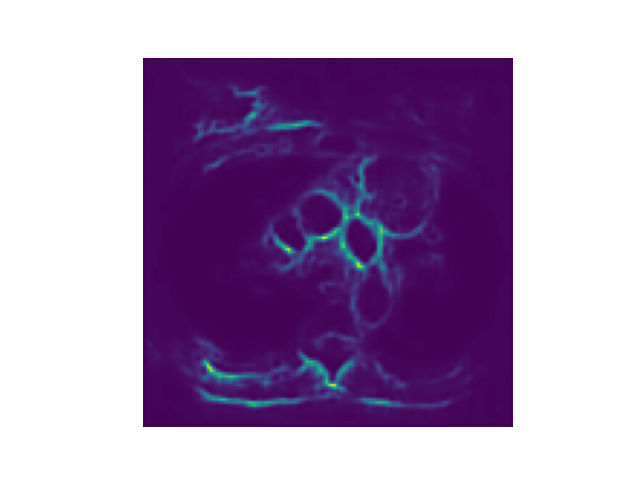}
\end{subfigure}

\caption{Results of the Fast Marching Energy CNN for images outside the scope of the training database.Top row: segmentation of outside the training scope. Bottom row: Potential output by the CNN before fast marching.}
\label{fig:tests}
\end{figure}

We further studied the properties of the generated potential of our CNN by testing it with dissimilar MRI images found randomly through an image search on Figure \ref{fig:tests} where activated areas correspond to the segmentation ranging from yellow to green for confidence. The results for the last two MRI images show that while the algorithm does not properly segment the tumour (as the predicted barycenter for initialization of the Fast Marching is not correctly placed), the learned filter detects small contours similar to tumours, focusing on the shape of the different objects.

\section{Conclusion}

Unlike traditional methods that focus solely on improving segmentation scores, our approach prioritizes the preservation of the tumour's geometrical structure. We have showed that it was possible to learn an interesting potential in order to segment brain tumours as unit balls of geodesic distances, and reach almost state-of-the-art performances on this task. By doing so, our method avoids the limitations of relying solely on convolutional operations, leading to more accurate and reliable results. Our approach offers an alternative path for tumour segmentation that considers both the quality of the segmentation and the preservation of the tumour's structure. This opens new possibilities in terms of geometrical and topological priors for all kinds of tasks.


Further works include extending this framework to general Riemannian metrics, evaluating the capabilities in terms of transfer learning of networks taught with our approach (see Figure \ref{fig:tests}) and another possible direct extension of our framework is to try and include multiple connected components for the segmentation mask generated, for which one needs to know or find the number of components in the mask.

\subsubsection{Acknowledgements} 
This work is in part supported by the French government under management of Agence Nationale de la Recherche as part of the "Investissements d’avenir" program, reference ANR-19-P3IA-0001 (PRAIRIE 3IA Institute).

\bibliographystyle{splncs04.bst}
\bibliography{bib}
\end{document}